\documentclass[runningheads]{llncs}
\usepackage[T1]{fontenc}
\usepackage{graphicx}
\usepackage{amsfonts,amssymb} 
\usepackage{amsmath}
\usepackage{multicol}
\usepackage{wasysym} 
\usepackage{pifont}  
\usepackage{amsfonts}
\usepackage{array}
\usepackage{booktabs}
\usepackage{float}
\usepackage{multirow}
\usepackage{color}
\usepackage{subfigure}

\usepackage{hyperref}
\usepackage{diagbox}

\usepackage{pifont} 

\newcommand{\Checkmark}{\ding{51}} 
\newcommand{\XSolidBrush}{\ding{55}}

\begin{document}
\title{Self-Prior Guided Mamba-UNet Networks for Medical Image Super-Resolution}
%
%
\author{Zexin Ji\inst{1,2,4} \and
Beiji Zou\inst{1,2} \and
Xiaoyan Kui\inst{1,2}* \and
Pierre Vera\inst{4} \and
Su Ruan\inst{3}
}
\authorrunning{Zexin Ji et al.}

\institute{School of Computer Science and Engineering, Central South University,
Changsha, 410083, China \and
Hunan Engineering Research Center of Machine Vision and Intelligent
Medicine, Central South University, Changsha, 410083, China \and
University of Rouen-Normandy, LITIS - QuantIF UR 4108, F-76000, Rouen, France \and
Department of Nuclear Medicine, Henri Becquerel Cancer Center, Rouen, France}
\maketitle              
\begin{abstract}

In this paper, we propose a self-prior guided Mamba-UNet network (SMamba-UNet) for medical image super-resolution. Existing methods are primarily based on convolutional neural networks (CNNs) or Transformers. CNNs-based methods fail to capture long-range dependencies, while Transformer-based approaches face heavy calculation challenges due to their quadratic computational complexity. Recently, State Space Models (SSMs) especially Mamba have emerged, capable of modeling long-range dependencies with linear computational complexity. Inspired by Mamba, our approach aims to learn the self-prior multi-scale contextual features under Mamba-UNet networks, which may help to super-resolve low-resolution medical images in an efficient way.
Specifically, we obtain self-priors by perturbing the brightness inpainting of the input image during network training, which can learn detailed texture and brightness information that is beneficial for super-resolution. Furthermore, we combine Mamba with Unet network to mine global features at different levels. We also design an improved 2D-Selective-Scan (ISS2D) module to divide image features into different directional sequences to learn long-range dependencies in multiple directions, and adaptively fuse sequence information to enhance super-resolved feature representation. Both qualitative and quantitative experimental results demonstrate that our approach outperforms current state-of-the-art methods on two public medical datasets: the IXI and fastMRI.
\end{abstract}

\section{INTRODUCTION}

Medical imaging techniques play a vital role in supporting clinical diagnoses. However, obtaining high-quality medical images demands high-quality medical imaging equipment and a longer imaging time. With the advancement of deep learning, super-resolution (SR) is considered as a promising direction for cost-saving improvements in image quality. Medical image super-resolution aims to restore the corresponding high-resolution images by adding missing details in low-resolution (LR) medical images.

To realize medical image super-resolution, initial research utilized methods of interpolation~\cite{1} and optimization~\cite{2}. These methods are simple to implement, but not precise enough in restoring image details. Subsequently, massive CNNs-based image super-resolution methods~\cite{8,DBLP:conf/miccai/LiuCWL19,DBLP:journals/cmpb/QiuZLZZ20,DBLP:journals/cmpb/QiuCW22} have come up to learn the nonlinear mapping relationship and representative features. It demonstrates superior performance in the field of medical imaging. However, they are prone to local matching difficulties. This occurrence is credited to the inductive bias of CNNs, which limits their ability to capture long-range dependencies. Compared to CNNs, Vision Transformers~\cite{DBLP:conf/iclr/DosovitskiyB0WZ21} have emerged for modeling non-local dependencies. Leveraging the ability of self-attention of Transformer, researchers have designed tailored approaches to achieve super-resolution~\cite{25,DBLP:conf/bibm/JiKLZLDZ23,DBLP:conf/miccai/ForiguaEA22,DBLP:journals/vc/HuangLTHWCS23}. However, Transformer-based methods face excessive computational complexity and memory requirements due to the attention calculation of the pairwise affinity. Moreover, these methods are still insufficient to fully mine the precise features hidden within complex medical image distributions. Mamba~\cite{DBLP:journals/corr/abs-2312-00752} is now widely applied as an emerging sequence model, originating from tasks in natural image processing. It has been proposed for learning visual representations that can efficiently capture long-distance dependencies of images.

Motivated by these insights, we propose a self-prior guided Mamba-UNet network for medical image super-resolution, which taps into the potential of lightweight long-range modeling and fully mining super-resolved features. To achieve this, we design a self-prior learning in a super-resolution network. Specifically, the input is disturbed by a brightness inpainting in the feature distribution during training, leading to an incompleteness of the input. This can enhance the ability of the model to mine the own super-resolved feature information. We also develop a Mamba-based Unet to learn the multi-scale contextual dependencies of images from hierarchical levels. Mamba can calculate long-range dependencies using linear complexity. Compared with traditional 2D-Selective-Scan in vision Mamba, we further design the improved 2D-Selective-Scan (ISS2D) model to adaptively integrate relevant information between image sequences from different directions, thereby alleviating the lack of causal reasoning in the image field.

The contributions of our approach can be summarized as follows:

1) We first build a Mamba network for medical image super-resolution. With the designed Mamba-based Unet network, we can simultaneously learn the long-range dependencies of features at different scales.

2) We design self-prior learning to improve the local texture and brightness refinement by borrowing self-reasoning features, which can significantly strengthen the generation of semantically coherent features.

3) We devise an improved 2D-Selective-Scan (ISS2D) module to dynamically model the correlation between features of four direction sequences in images. It can better fuse four direction weighted features for image super-resolution. 

4) Qualitative and quantitative experiment results on IXI and fastMRI datasets demonstrate that our approach achieves superior performance comparing to the existing state-of-the-art super-resolution methods.

\section{RELATED WORK}

\subsection{Medical Image Super-Resolution}
The basic CNNs have been proposed to deal with super-resolution. SRCNN~\cite{8} first introduced deep CNNs into super-resolution and achieved promising results. Then, many improved methods based on SRCNN have emerged, and significant progress has been made in medical image super-resolution tasks~\cite{DBLP:journals/cmpb/QiuZLZZ20,DBLP:conf/miccai/LiuCWL19,DBLP:journals/cmpb/QiuCW22,DBLP:journals/inffus/UmirzakovaAKW24}. Specifically, Qiu \emph{et al.} proposed EMISR~\cite{DBLP:journals/cmpb/QiuZLZZ20} based on the improvement of SRCNN for the knee MR image super-resolution. Liu \emph{et al.}~\cite{DBLP:conf/miccai/LiuCWL19} developed an edge-enhanced super-resolution generative adversarial networks (EE-SRGAN) for medical image super-resolution. A dual U-Net residual network (DURN) ~\cite{DBLP:journals/cmpb/QiuCW22} was designed to enhance cardiac MR image resolution. These CNNs-based approaches primarily capture local patterns of images due to the nature of convolution operations, lacking of ability to model long-range dependencies. Thanks to the long-range representation ability of self-attention in Transformer, it has shown superior performance compared to CNNs-based methods. Dosovitskity \emph{et al.}~\cite{DBLP:conf/iclr/DosovitskiyB0WZ21} firstly proposed Vision Transformer (ViT) for computer vision task, such as medical image classification~\cite{DBLP:journals/tmi/ShiTGLWGLF23} and segmentation~\cite{DBLP:journals/tmi/JiC24} tasks. The same holds in the area of medical image super-resolution~\cite{25,DBLP:conf/miccai/ForiguaEA22,DBLP:conf/bibm/JiKLZLDZ23,DBLP:journals/vc/HuangLTHWCS23}. SuperFormer~\cite{DBLP:conf/miccai/ForiguaEA22} explored the effectiveness of a swim transformer in the 3D MR image super-resolution. The TransMRSR~\cite{DBLP:journals/vc/HuangLTHWCS23} further combined the local information of CNNs and the global information of Transformers for medical image super-resolution. Although these Transformer-based methods have achieved superior results, they require a large amount of GPU memory to train the model.

\subsection{State Space Models}
Recently, State Space Models (SSMs)~\cite{DBLP:conf/nips/GuJGSDRR21}, with roots in classical control theory, have made their way into deep learning, showcasing potential as an effective architecture for modeling sequences. These models blend the characteristics of recurrent neural networks (RNNs) and CNNs, establishing a novel approach to sequence analysis. Compared with Transformers with large parameter scales, SSMs have good characteristics of linear scaling of sequence length and modeling long-range dependencies. The Structured State Space Sequence Model named as S4~\cite{DBLP:conf/iclr/GuGR22} is a pioneering work in deep state space models. It models in a content-agnostic static representation. Recently, a new SSM architecture named Mamba~\cite{DBLP:journals/corr/abs-2312-00752} was proposed based on S4. The authors designed a selective scan space state sequential model (S6) that is a data-related SSM with selective mechanisms and efficient hardware design. The breakthrough of Mamba in sequence modeling has aroused great interest in the field of computer vision~\cite{zhu2024vision}. Some works have applied Mamba as a sequence model backbone to visual tasks, such as medical image segmentation~\cite{xing2024segmamba,DBLP:journals/corr/abs-2402-02491}, video understanding~\cite{DBLP:conf/cvpr/WangZWYLOH23},  classification~\cite{DBLP:journals/corr/abs-2401-10166} and video understanding~\cite{DBLP:conf/cvpr/WangZWYLOH23}. In this work, we first explore the potential of Mamba for medical image super-resolution.
\begin{figure}[t]
    \centering
    \includegraphics[width=0.5\linewidth]{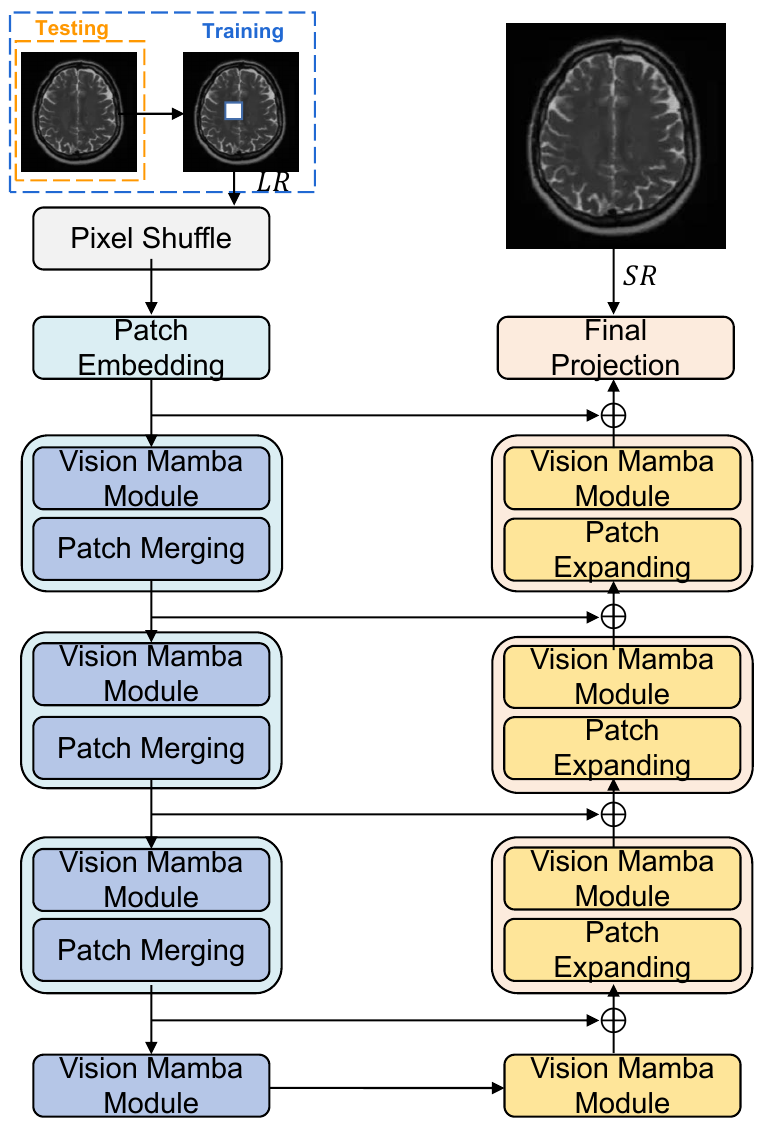}
    \caption{Our SMamba-Unet framework primarily comprising a patch embedding layer, a Mamba-based encoder module, a Mamba-based decoder module, and a final projection layer.}
    \label{fig:Framework1}
\end{figure}

\section{METHOD}

\subsection{Overview}

The core structure of our SMamba-Unet is depicted in Figure \ref{fig:Framework1}. Before being fed into the network, we perturb the image based on a central random brightness inpainting, which enables the network to mine its own prior information for image super-resolution. Then, we employ a pixel shuffle technique for learnable upscaling of low-resolution images, aiming to preserve the integrity of image information. 
Traditional upscaling methods like bicubic interpolation enlarge image pixels, often losing sharpness of organ contours and texture detail. In contrast, the pixel shuffle technique in neural networks rearranges multiple channel outputs into a higher-resolution image, shuffling pixels into the correct spatial order to better recover contours and textures.
Our method primarily comprises patch embedding layer, a Mamba-based encoder module, a Mamba-based decoder module, and a final projection layer. In detail, the input image is initially split into non-overlapping patches via patch embedding layer. The Mamba-based encoder mainly contains four layers. Each layer is equipped with the vision Mamba module and patch merging, which can extract contextual features and further increase the number of channels, respectively. Within the Mamba-based decoder, each stage progressively merges information from skip connections to upscale features while simultaneously decreasing the channel count by the vision Mamba module and patch expanding. Finally, the super-resolved medical image is generated by the final projection layer.

\subsection{Self-prior Learning}
From the error map shown in Figure \ref{fig:2-Scale-IXI-fastMRI} and Figure \ref{fig:4-Scale-IXI-fastMRI}, it can be seen that the main difference between the super-resolved image and the high-quality image lies in the texture areas with significant brightness changes. 
It has been confirmed that the repetitive characteristics of an image serve as an effective prior for image super-resolution~\cite{DBLP:journals/pami/SuGCYC23}. 
Therefore, we propose self-prior learning to mine 
valuable repetitive internal examples of super-resolved images to enhance the learning of texture and brightness feature representations in perturbed images. 
Unlike natural scenes, medical images contain a large amount of invalid noise background, while physicians mainly focus on the foreground areas with rich information in the middle of medical images where the organs of interests are presented. Therefore, we first define the width and height of the area of interest in the image, and the starting coordinates of the area in the image. Then, we randomly select a position within the region of interest to add a 5$\times5$ brightness block to perturb the image, and enhance the learning ability of texture and brightness features in the image through self-reasoning inpainting. During our testing, we input the original low-resolution medical images to obtain high-quality images.

\subsection{Vision Mamba Module}

Before introducing Mamba~\cite{DBLP:journals/corr/abs-2312-00752}, let's first revisit the Transformer. The Transformer views any text input as a sequence of tokens. It creates a self-attention matrix that compares each token with every other token. The values in the matrix encode the correlations between them. Generating a self-attention matrix for a sequence of length L requires approximately $L^2$ computations, which is quite computationally intensive. Currently, Mamba has emerged as a promising approach, which can be trained in parallel, while still performing inference that scales linearly with the length of the sequence. It is utilized to characterize the state representations and to forecast their subsequent states based on given inputs.
It transforms a 1-D function or sequence $x(t) \in \mathbb{R}$ into the output $y(t) \in \mathbb{R}$ via a hidden state $h(t) \in \mathbb{R}^{\mathbb{N}}$, which usually realizes through linear ordinary differential equations (ODEs).

\begin{equation}
    h^{\prime}(t)=\mathbf{A} h(t)+\mathbf{B} x(t), y(t)=\mathbf{C} h(t),
\end{equation}
where $\mathbf{A} \in \mathbb{R}^{\mathrm{N} \times \mathrm{N}}$ is the state matrix. $\mathbf{B} \in \mathbb{R}^{\mathbb{N} \times 1}$ and $ \mathbf{C} \in \mathbb{R}^{1 \times \mathbb{N}}$ are the projection parameters.

The zero-order hold (ZOH) technique~\cite{DBLP:journals/tcas/GaliasY08} is used for discretizing ODEs into discrete functions, making it better adapted for deep learning contexts. By incorporating a timescale parameter $\Delta$, it facilitates the conversion of the continuous-time system matrices $\mathbf{A}$ and $\mathbf{B}$ into their discrete equivalents, noted as $\overline{\mathbf{A}}$ and $\overline{\mathbf{B}}$. The specific steps involved in this conversion are meticulously designed to preserve the integrity of the original system's dynamics while making them compatible with the discrete computational environment of deep learning models. The discretization process is implemented as follows:
\begin{equation}\label{eq2}
\overline{\mathbf{A}}=\exp (\Delta \mathbf{A}), \overline{\mathbf{B}}=(\Delta \mathbf{A})^{-1}(\exp (\Delta \mathbf{A})-\mathbf{I}) \cdot \Delta \mathbf{B} .
\end{equation}

Following discretization, Equation (1) adopts a form suitable for discrete-time processing, as follows:

\begin{equation}\label{eq3}
h_t=\overline{\mathbf{A}} h_{t-1}+\overline{\mathbf{B}} x_t, y_t=\mathbf{C} h_t .
\end{equation}
\begin{figure}[t]
    \centering
    \includegraphics[width=0.8\linewidth]{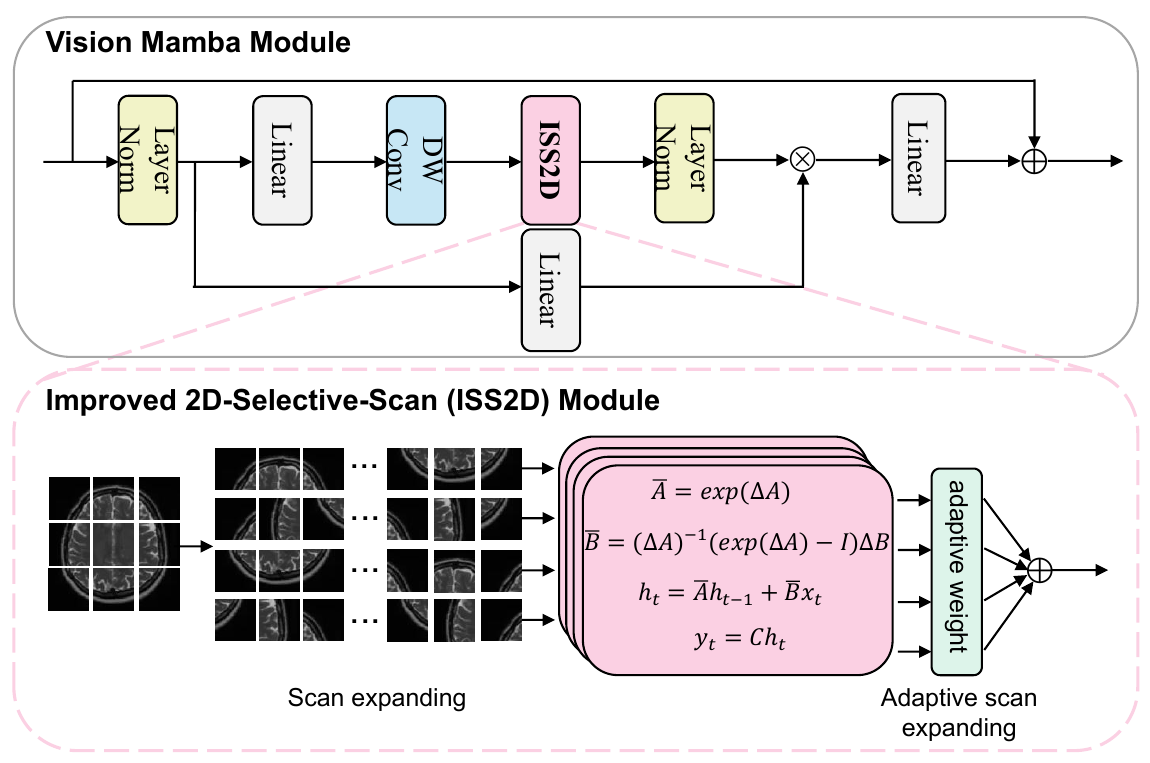}
    \caption{The framework of the vision Mamba module and improved 2D selective scan (ISS2D) module.}
    \label{fig:Framework2}
\end{figure}

Mamba further presents a groundbreaking method within the realm of SSMs through its introduction of Selective Structured State Space Sequence Models (S6). This advancement permits dynamic parameterization of the SSMs, where the parameters $\overline{\mathbf{B}}$, $\mathbf{C}$  and $\Delta$ are directly determined by the input data, facilitating a model adaptation unique to each input. Defined by its linear complexity and further optimized for hardware efficiency, Mamba stands out for its exceptional capability in handling the modeling of lengthy sequences.

Similar to the Transformer, Mamba also processes its input by dividing the image into a series of image patches. Figure \ref{fig:Framework2} shows the specific structure of the vision Mamba module. Initially, the input patch feature map undergoes layer normalization, after which it splits into two distinct paths. The initial path processes the input via a linear layer. Concurrently, in the alternate path, the input is subjected to a linear layer, then it proceeds through a depthwise separable convolution (DW Conv), then advancing into the improved 2D-Selective-Scan (ISS2D) and layer normalization. Integration of these divergent pathways is achieved through a multiplication process and a linear layer, combining the processed outputs with the initial input to further enhance feature representation.

\subsection{Improved 2D-Selective-Scan (ISS2D)}

The detailed implementation of improved 2D-Selective-Scan (ISS2D) module is illustrated in Figure \ref{fig:Framework2}. Unlike sentences in natural language processing, the patches of the image lack a direct inferential relationship. Therefore, we use scan expanding directions corresponding to the horizontal and vertical sequences of medical images, that is, scanning from left to right and from right to left, coupled with scanning from top to bottom and from bottom to top.  Different sequences are further processed by equations we mentioned before. 
The existing 2D-Selective-Scan Mamba models simply add the scanning results from all directions with equal weights to obtain output features, overlooking the importance of different sequential directions for medical image super-resolution.
Therefore, we propose the ISS2D module to blend the four sequences with the weighting coefficients which are learned automatically, making the fusion of the four directions more relevant. 
Each directional sequence has a coefficient corresponding to its contribution. Summarizing features with the weight coefficients from four different directions can better enhance the global spatial feature connections of medical images across different directions.

\subsection{Loss Function}

We use the $\mathcal{L}_{1}$ loss to compute the absolute value of pixel differences between ground truth $HR$ and super-resolved image $SR$, expressed as follows:
\begin{equation}
	\begin{aligned}
		\mathcal{L}_{1}(SR, HR)=\frac{1}{n}\sum_{i=1}^{n}\left|SR_i-HR_i\right|,
	\end{aligned}
\end{equation}
where $n$ is the number of images. $\mathcal{L}_{1}$ loss as pixel-level loss, leading to overly smooth generated images that miss out on semantic details. Therefore, we also use perceptual loss \cite{DBLP:conf/eccv/JohnsonAF16} to measure the discrepancy between images at feature representation levels, thereby improving the visual appeal of super-resolved medical images. Its calculation is as follows:
\begin{equation}
	\mathcal{L}_{\text {perceptual}}^{\phi, j}(SR, HR)=\frac{1}{n}  \sqrt{\sum_{i=1}^{n}\left(\phi_{j}(SR_i)-\phi_{j}(HR_i)\right)^{2}},
\end{equation}
where $\phi$ represents the VGG19 network pre-trained on ImageNet.
Through training on millions of labeled images, the VGG19 network has learned the ability to recognize a wide range of visual patterns. Therefore, we use it to help our feature extraction.
The $\phi_{j}(\cdot)$ corresponds to the feature map produced by the $j^{th}$ layer of the network $\phi$. $j$ is 36.

The final loss function is depicted as follows:

\begin{equation}
\begin{split}
Loss = \mathcal{L}_{1} + \beta \mathcal{L}_{\text {perceptual}}^{\phi, j}, 
\end{split}
\end{equation}
where $\beta$ is the weighting factor.

\section{EXPERIMENTS}

\subsection{Datasets and Metrics}
We assessed our method on T2-weighted MRI brain and knee images from the IXI\footnote{http://brain-development.org/ixi-dataset/} and fastMRI\footnote{https://fastmri.med.nyu.edu/} dataset, respectively. The slices in the IXI dataset have a fixed size of 256 $\times$ 256 pixels with a resolution of 1mm, while the fastMRI dataset consists of slices having size of 320$\times$320 pixels with a resolution 0.5mm. In our experiment, we used 368 subjects from the IXI dataset for the training and reserved 92 subjects for testing. Regarding the fastMRI dataset, we utilized the data for training on 227 subjects and for conducting tests on 45 subjects. We utilized the peak signal-to-noise ratio (PSNR) and structural similarity index (SSIM)~\cite{45} commonly used in super-resolution to quantitatively evaluate the medical image quality. 

\begin{table}[b]
\centering
\caption{Quantitative results with different methods on fastMRI and IXI dataset under 2$\times$ upsampling factor.}\label{tab1}
\setlength{\tabcolsep}{5mm}{
\begin{tabular}{c|cc|cc}
\hline
\multirow{2}{*}{Method} & \multicolumn{2}{c|}{fastMRI 2$\times$} & \multicolumn{2}{c}{IXI 2$\times$} \\  
                        & PSNR$\uparrow$           & SSIM$\uparrow$           & PSNR$\uparrow$          & SSIM$\uparrow$        \\ \hline
SRCNN                   & 25.82          & 0.5602       & 29.23        & 0.8649     \\
VDSR                    & 27.42          & 0.6263        & 29.79        & 0.8772     \\
FMISR                   & 26.19          & 0.5583        & 29.50      & 0.8685     \\
T$^{2}$Net                   & {32.00}           & 0.7158         & 31.31      & 0.9035     \\
DiVANet                     & \underline{31.98}       & \underline{0.7169}        & \underline{33.15}      & \underline{0.9320}      \\
SMamba-Unet(ours)                     & \textbf{32.06}        & \textbf{0.7180}        & \textbf{33.36}       & \textbf{0.9355}      \\ \hline
\end{tabular}}
\end{table}

\begin{table}[t]
\centering
\caption{Quantitative results with different methods on fastMRI and IXI dataset under 4$\times$ upsampling factor.}\label{tab22}
\setlength{\tabcolsep}{5mm}{
\begin{tabular}{c|cc|cc}
\hline
\multirow{2}{*}{Method} & \multicolumn{2}{c|}{fastMRI 4$\times$} & \multicolumn{2}{c}{IXI 4$\times$} \\  
                        & PSNR$\uparrow$           & SSIM$\uparrow$           & PSNR$\uparrow$          & SSIM$\uparrow$        \\ \hline
SRCNN                   & 19.74          & 0.3653       & 28.12        & 0.8357     \\
VDSR                    & 20.31          & 0.3839        & 28.34        & 0.8392     \\
FMISR                   & 24.35          & 0.5207        & 28.27      & 0.8349     \\
T$^{2}$Net                   & {30.56}           & {0.6244}         & 29.73      & 0.8773     \\
DiVANet                     & \underline{30.62 }       & \underline{0.6352}        & \underline{30.46}      & \underline{0.8946}      \\
SMamba-Unet(ours)                       & \textbf{30.70}        & \textbf{0.6361}        & \textbf{31.13}       & \textbf{0.9081}      \\ \hline
\end{tabular}}
\end{table}

\begin{figure}[t]
    \centering
    \includegraphics[width=1\linewidth]{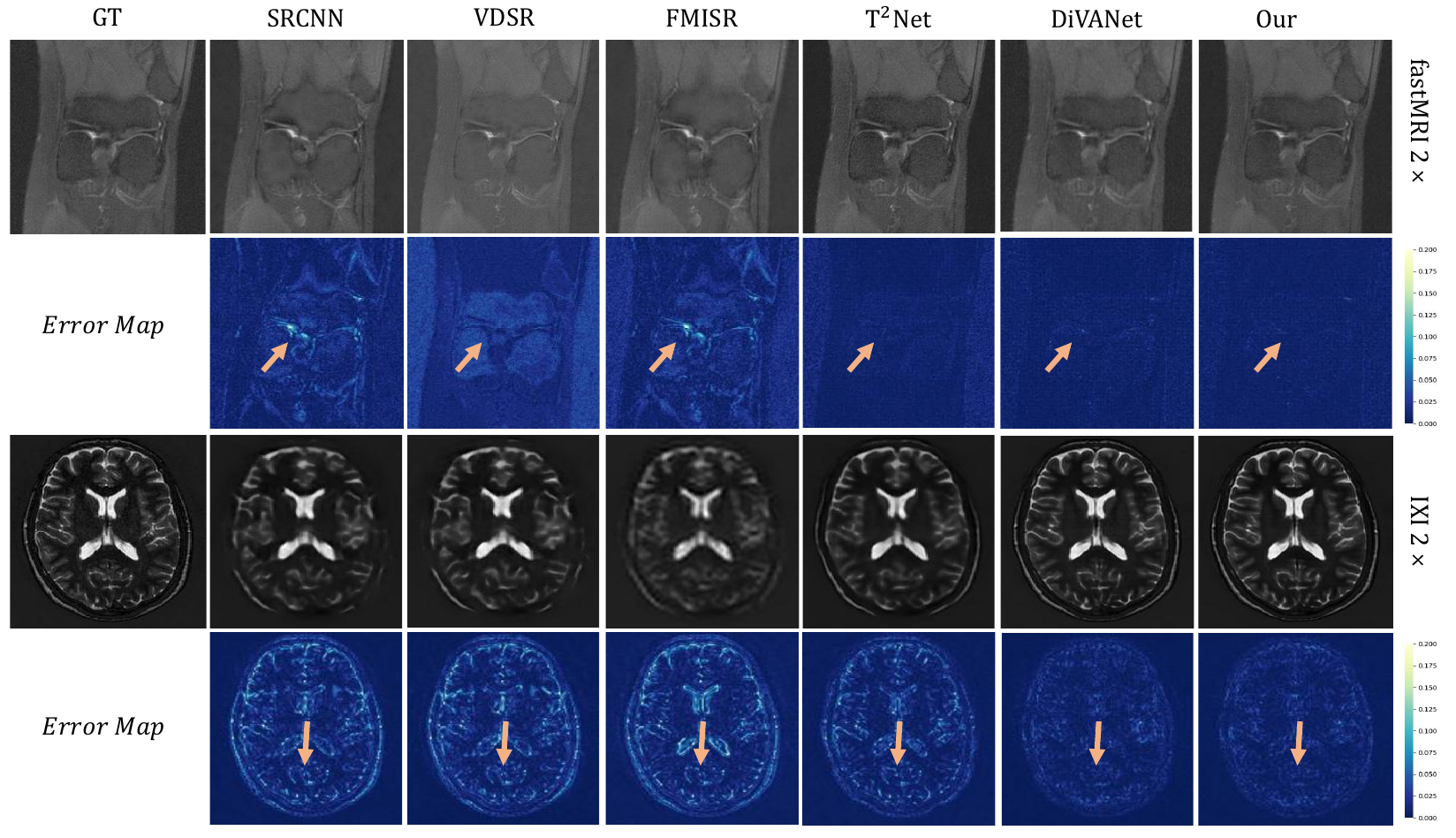}
    \caption{Qualitative results on fastMRI and IXI dataset under 2$\times$ upsampling factor. The significant differences between different methods are shown by the yellow arrow.}
    \label{fig:2-Scale-IXI-fastMRI}
\end{figure}

\subsection{Implementation Details} 

To obtain low-resolution medical images for 2$\times$ and 4$\times$ super-resolution, we initially apply the degradation model in the frequency domain~\cite{36} to generate low-resolution images that better match the distribution of real scenes.
Our proposed approach was implemented using the PyTorch framework on a NVIDIA RTX A6000 GPU. We used Adam optimizer to train the network with the initial learning rate of $1 \times 10^{-4}$. We incorporated 4 vision Mamba blocks in each level, with the channel count [96,128,384,768] for each level, respectively. The dropout rate used within the vision Mamba is 0.3. 
The dimension of state vectors is 16. The weighting factor $\beta$ is 0.01.

\subsection{Comparison with the State-of-the-arts}

\textbf{Quantitative Comparison.}
We compared our approach with the
SRCNN~\cite{8}, VDSR~\cite{9}, FMISR~\cite{19},
T$^2$Net~\cite{12}, and DiVANet~\cite{DBLP:journals/pr/BehjatiRFHMG23} on the IXI and fastMRI dataset under $2 \times$ and $4 \times$ upsampling factors.
Table \ref{tab1} and Table \ref{tab22} show the quantitative comparison results. It can be seen from the table that our approach achieved the highest PSNR and SSIM scores compared with other methods for all scaling factors. The primary reasons are: i) the proposed self-prior learning can let the model mine self-exemplar in the original image, fully investigating its own texture and brightness information and enhancing feature representation capability; and ii) 
the designed Mamba-based Unet module not only can exhaustively exploit local features from different levels but also can fully explore the long-range dependencies of features at different scales.

\begin{figure}[t]
    \centering
    \includegraphics[width=1\linewidth]{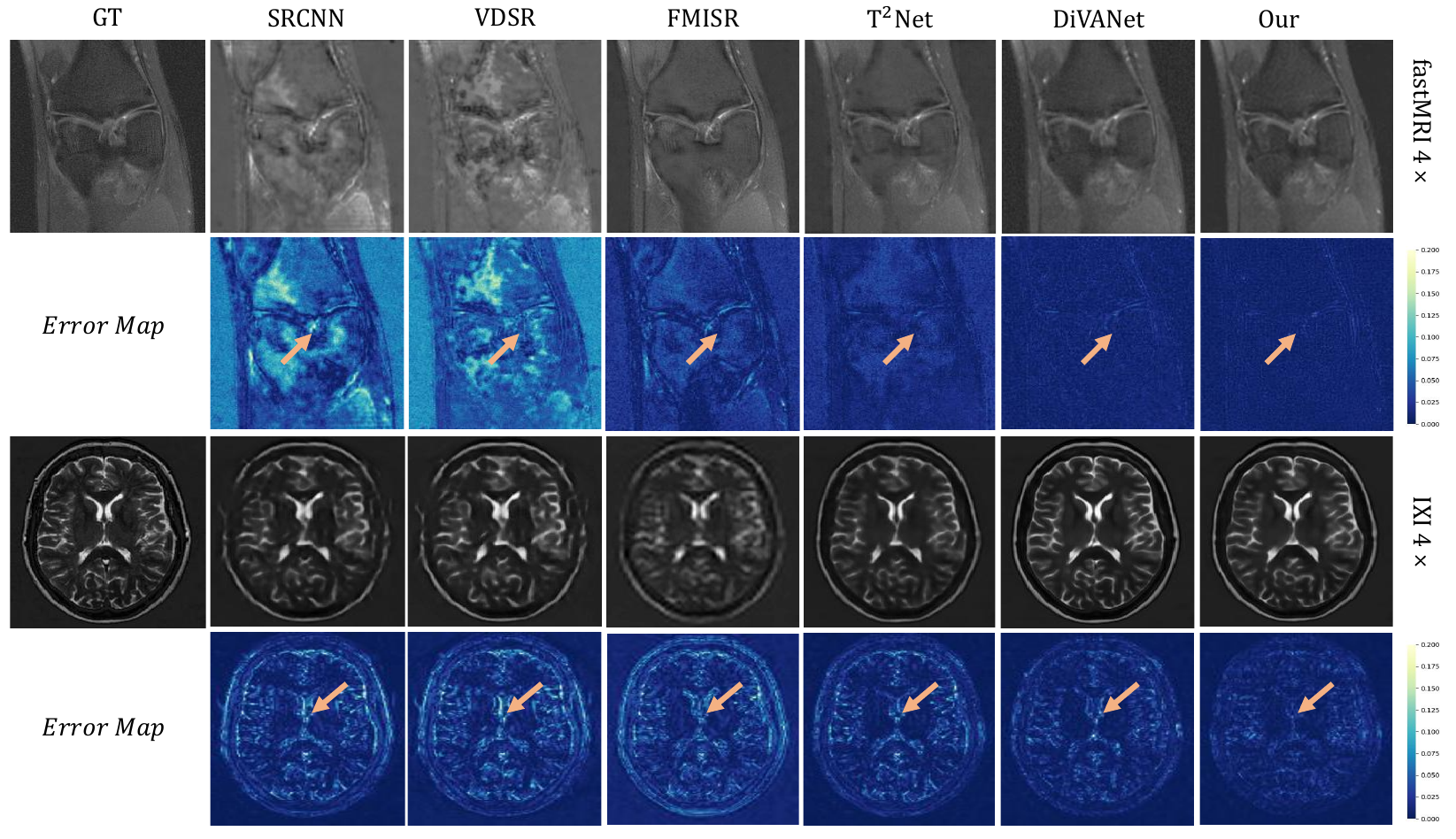}
    \caption{Qualitative results on fastMRI and IXI dataset under 4$\times$ upsampling factor. The significant differences between different methods are shown by the yellow arrow.}
    \label{fig:4-Scale-IXI-fastMRI}
\end{figure}

\textbf{Qualitative Comparison.}
The qualitative experimental results are shown in Figure \ref{fig:2-Scale-IXI-fastMRI} and Figure \ref{fig:4-Scale-IXI-fastMRI}. 
In addition to displaying super-resolved images, we also show the corresponding error map to better illustrate the differences with ground truth in detail. The darker the error map, the better the super-resolved image. The specific difference between medical images generated by different methods is indicated by yellow arrows.
It is evident that the image generated by SRCNN is very blurred. T$^2$Net and DiVANet can generate relatively clear soft tissue structures, although there are still some unpleasant blurry edges. From the overall error map, it can be seen that errors mainly exist in areas with significant changes in brightness. The designed self-prior learning method randomly uses brightness inpainting to complete damaged images in the foreground area. It can enhance the ability of the network to explore lost features and changes in brightness through its own information, thereby improving the learning capability of super-resolved features. Therefore, our approach has better overall clarity than other methods and can generate more delicate image details. This aligns with the findings from the quantitative analysis.

\subsection{Ablation Analysis}
\textbf{Effectiveness analysis of the key component.}
To explore the effectiveness of the key component in our SMamba-Unet, we conducted the ablation experiment using the IXI dataset under $2 \times$ upsampling factor. We use the pure Mamba-based Unet as baseline, then gradually add improved 2D-Selective-Scan (ISS2D) and self-prior learning (SPL) to it. From Table \ref{tab3}, it can be seen that the performance of the super-resolution network gradually improves as different components participate. The PSNR value increased from 33.18 to 33.36. The value of SSIM increased from 0.9245 to 0.9355.
This fully demonstrates that focusing on the importance of different sequence features of images can make the features more effective and accurate. In addition, self-prior learning makes the network pay more attention to its own texture and brightness feature information, which helps in the generation of super-resolved images.

\begin{table*}[t]
	\centering
	\setlength{\belowcaptionskip}{0.2cm}
	\caption{Ablation study with different components under $2 \times$ upsampling factor.}\label{tab3}
	\renewcommand\arraystretch{1.5} 
	\setlength{\tabcolsep}{2mm}{
		\label{tab2}
		\begin{tabular}{l|l|l|c|l|l}
			\hline
			Method  & Mamba & ISS2D & SPL & PSNR$\uparrow$    & SSIM$\uparrow$      \\ \hline
			\textit{Base} & \Checkmark  & \XSolidBrush   & \XSolidBrush    & 33.18 & 0.9245  \\ \hline
			\textit{Base+ISS2D}   & \Checkmark  & \Checkmark   & \XSolidBrush    & 33.26 & 0.9284  \\ \hline
			\textit{Base+ISS2D+SPL (SMamba-UNet)}  & \Checkmark  & \Checkmark   & \Checkmark    & \textbf{33.36} & \textbf{0.9355}  \\ \hline
	\end{tabular}}
\end{table*}

\begin{figure}[t]
    \centering
    \includegraphics[width=1\linewidth]{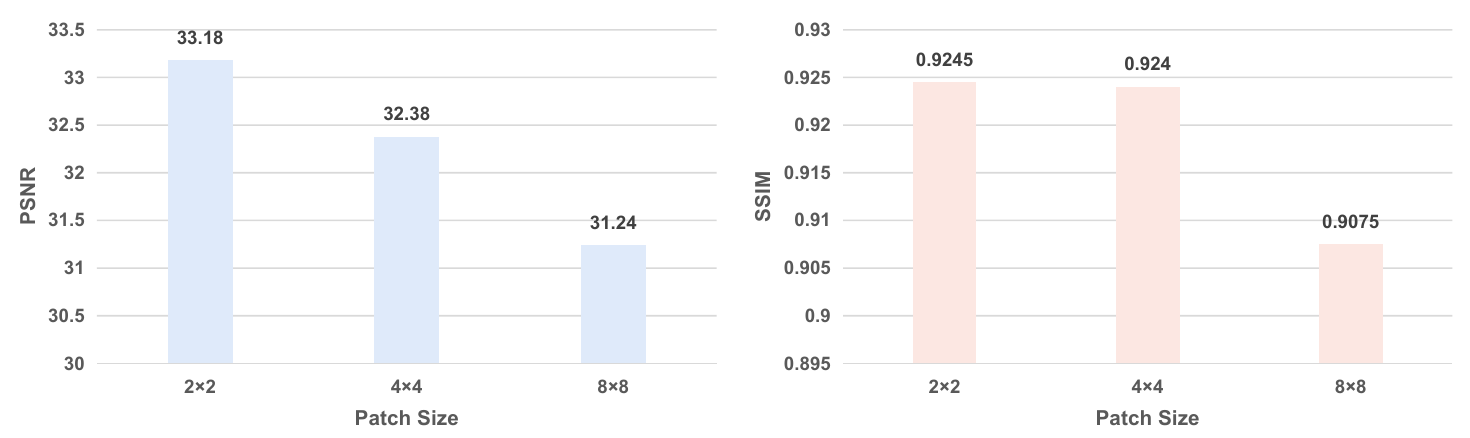}
    \caption{Ablation study with different patch sizes in vision Mamba module.}
    \label{fig:Ablation}
\end{figure}

\begin{table*}[t]
	\centering
	\setlength{\belowcaptionskip}{0.2cm}
	\caption{Ablation study with MambaUnet and TransUnet on model complexity. }
	\renewcommand\arraystretch{1.5} 
	\setlength{\tabcolsep}{1.0mm}{
		\label{tab4}
		\begin{tabular}{l|l|l|l|l|l}
			\hline
			Method   & Params[M]    & FLOPS[G] & GPU Memory[Mib]  & PSNR$\uparrow$ & SSIM$\uparrow$ \\ \hline
			MambaUnet   & 27.57 & 18.47& 4483 &  33.18&0.9245 \\ \hline
			TransUnet & 41.05 & 32.36 & 30579 &32.14 &0.9203 \\ \hline
	\end{tabular}}
\end{table*}

\noindent\textbf{Effectiveness analysis of patch size.}
In medical image super-resolution tasks using Mamba, selecting the appropriate patch size is an important design decision. To demonstrate the impact of different patch sizes on the performance of Mamba model, we set it to 2, 4, and 8, respectively, to observe the differences in performance. Figure \ref{fig:Ablation} shows that smaller patch sizes achieve better performance. A smaller patch size means that the Mamba will extract finer-grained features from the medical image, which can capture more detailed information. To balance performance and computational cost, we choose a patch size of 2 in our approach.

\begin{table}[t]
	\centering
	
	\setlength{\belowcaptionskip}{0.2cm}
	\caption{Ablation study on the weight $\beta$ in the loss function.  }
	\renewcommand\arraystretch{1.2} 
	\label{tab5}
	\setlength{\tabcolsep}{13mm}{
		\begin{tabular}{l|l|l}
			\hline
			$\beta$ &  PSNR$\uparrow$                                    & SSIM$\uparrow$                                             \\ \hline
			
			{0.1}                      & {33.28}                                 & {0.9347}                                           \\ \hline
			\textbf{0.01}                        & \textbf{33.36 }                                & \textbf{0.9355}                                           \\ \hline
			{0.001}       &{33.34} & {0.9350}  \\ \hline
			
	\end{tabular}}
\end{table}

\noindent\textbf{Effectiveness analysis of the baseline selection.}
We conducted baseline selection experiments to validate the effectiveness of methods based on Vision Mamba and Vision Transformer. We combine the Mamba and Unet as MambaUnet and combine the Transformer and Unet as TransUnet. Vision Transformers have high computational complexity when processing images because the computational complexity of the self-attention mechanism increases quadratically with the number of image patches. While, the hardware-aware algorithm of Mamba processes data with a linear relationship to the sequence length Table \ref{tab4} shows the Params, FLOPs, and GPU Memory of TransUnet and MambaUnet for 128 $\times$ 128 inputs on the IXI dataset.
It can be seen from the table that MambaUnet can achieve better performance with less computational complexity. This indicates that exploring Vision Mamba is an excellent approach to achieving efficient super-resolution.

\noindent\textbf{Effectiveness analysis of weight $\beta$ in the loss function:}
We evaluated the impact of weight $\beta$ in the loss function. Specifically, we conducted ablation studies by setting $\beta$ to $\left\{ 0.1,0.01,0.001 \right\}$. Table \ref{tab5} presents the quantitative evaluation results. The results indicate that $\beta = 0.01$ achieves the best performance than other settings.

\section{CONCLUSION AND FUTURE WORK}

In this paper, we have developed a self-prior guided Mamba-UNet network (SMamba-UNet) for medical image super-resolution. Specifically, our method learns to exploit self-prior multi-scale contextual features within Mamba-UNet networks, potentially facilitating efficient medical image super-resolution. The quantitative and qualitative performance of our SMamba-UNet on the IXI and fastMRI datasets confirms the effectiveness of our approach. How to apply vision Mamba for the multi-model medical image super-resolution, which is an intriguing future path of our approach.

\subsubsection{Acknowledgements} The work was supported by the National Key R$\&$D Program of China (No.2018AAA0102100); the National Natural Science Foundation of China (Nos.U22A2034, 62177047); the Key Research and Development Program of Hunan Province (No.2022SK2054); High Caliber Foreign Experts Introduction Plan funded by MOST; Central South University Research Programme of Advanced Interdisciplinary Studies (No.2023QYJC020); China
Scholarship Council (No.202306370195).

%
%
%
%

\bibliographystyle{splncs04}
\bibliography{mybibfile}

\end{document}